\documentclass{ifacconf}

\usepackage{graphicx}      
\usepackage[round]{natbib}        
\usepackage{url}
\usepackage{amsmath}
\usepackage{amssymb}
\usepackage{xcolor}
\usepackage{subfigure}
\usepackage{enumitem}
\usepackage{float}

\newtheorem{assumption}{Assumption}
\begin{document}
\begin{frontmatter}

\title{Tiny Learning-Based MPC for Multirotors: Solver-Aware Learning for Efficient Embedded Predictive Control\thanksref{footnoteinfo}} 

\thanks[footnoteinfo]{This research was supported by the Natural Sciences and Engineering Research Council of Canada.}

\author[First]{Babak Akbari} 
\author[First]{Justin Frank} 
\author[First]{Melissa Greeff}

\address[First]{Robora Lab, Queen's University, Kingston, ON, Canada. (e-mails: \{babak.akbari, 21jtf4, melissa.greeff\}@queensu.ca).}

\begin{abstract}   Tiny aerial robots hold great promise for applications such as environmental monitoring and search-and-rescue, yet face significant control challenges due to limited onboard computing power and nonlinear dynamics. Model Predictive Control (MPC) enables agile trajectory tracking and constraint handling but depends on an accurate dynamics model. While existing Learning-Based (LB) MPC methods, such as Gaussian Process (GP) MPC, enhance performance by learning residual dynamics, their high computational cost restricts onboard deployment on tiny robots. This paper introduces Tiny LB MPC, a co-designed MPC framework and optimization solver for resource-constrained micro multirotor platforms. The proposed approach achieves 100 Hz control on a Crazyflie~2.1 equipped with a Teensy~4.0 microcontroller, demonstrating a 43\% average improvement in tracking performance over existing embedded MPC methods under model uncertainty, and achieving the first onboard implementation of LB MPC on a 53~g multirotor.
\end{abstract}


\end{frontmatter}

\section{Introduction}

Tiny, low-cost aerial robots hold great potential for applications such as environmental monitoring~\citep{yang_photochemically_2024}, agriculture~\citep{jafferis_untethered_2019}, and search-and-rescue~\citep{zhou_swarm_2022}, owing to their compact size and deployment flexibility. However, these missions demand precise control in challenging environments while operating under severe onboard computational constraints. Achieving agile, autonomous flight for tiny aerial robots—see~\citep{cai_survey_2014} for definition—requires generating and executing feasible trajectories near the physical limits of the vehicle despite nonlinear dynamics, modeling errors, and environmental disturbances such as drag or ground effect.
\begin{figure}
    \centering
\includegraphics[width=0.48\textwidth]{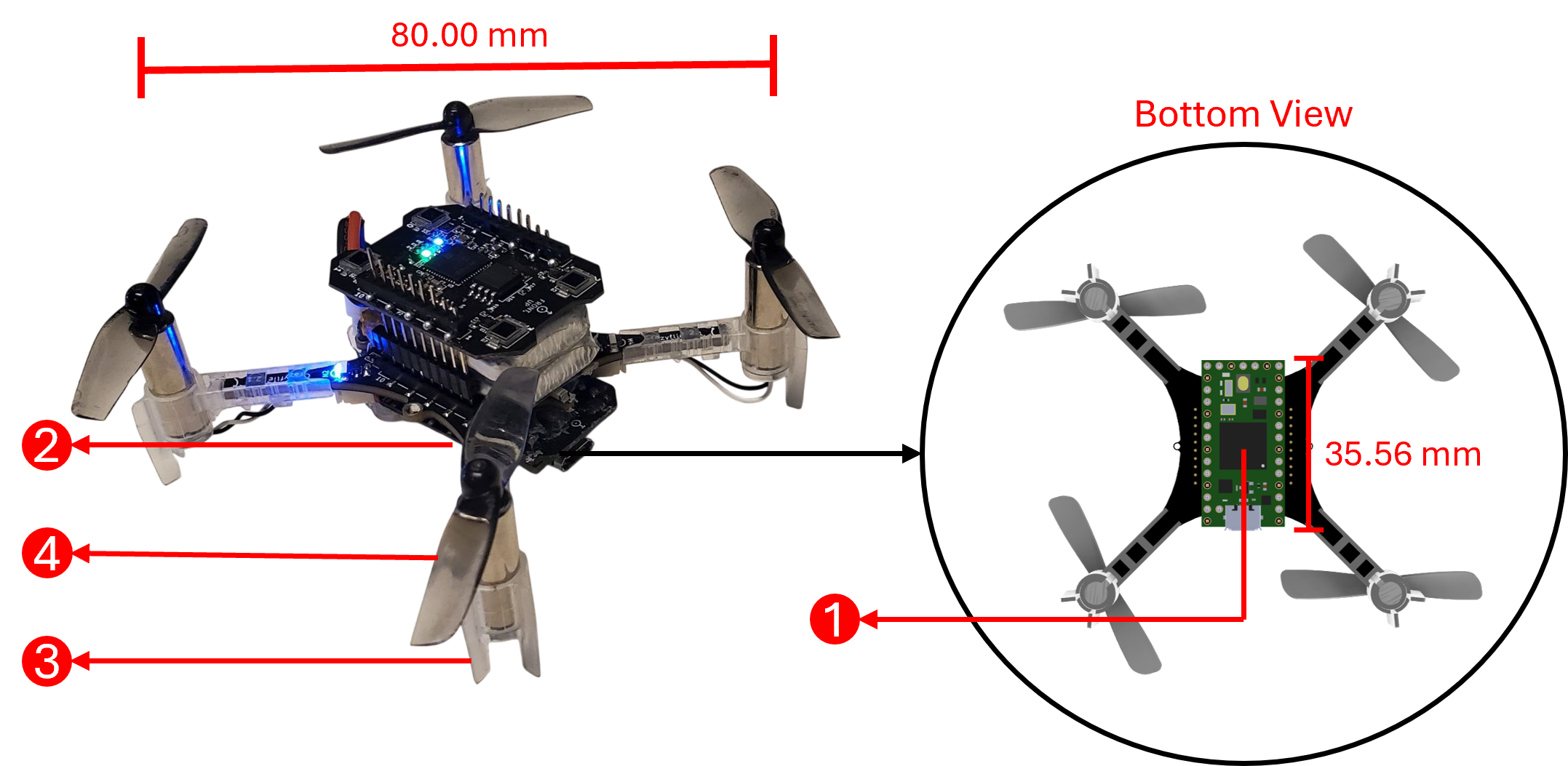}
    \vspace{-0.3cm}
    \caption{Hardware setup for the proposed Tiny Learning-Based MPC framework. 
The controller runs entirely onboard a \textbf{(1) Teensy 4.0} microcontroller 
(600~MHz, 1~MB RAM) mounted on a \textbf{(2) custom expansion board} and interfaced 
with a \textbf{(3) Crazyflie~2.1} platform. The complete system, including 
\textbf{(4) upgraded motors and propellers} (maximum collective thrust 
0.764~N), weighs 53~g in total. The setup demonstrates full onboard 
learning-based predictive control on a resource-constrained multirotor.} 
    \label{fig:teensy}
\end{figure}
Model-free reinforcement learning can operate on microcontrollers (e.g., RLtools~\citep{eschmann_rltools_2024}), but typically lacks explicit constraint handling, safety guarantees, and requires extensive training data. Model Predictive Control (MPC), in contrast, offers high tracking performance while explicitly enforcing state and input constraints~\citep{kamel_linear_2017}. Nonlinear MPC (NMPC) has been successfully demonstrated on full-scale multirotors for agile flight~\citep{hanover_performance_2022, tzoumanikas_nonlinear_2020, bicego_nonlinear_2020, nan_nonlinear_2022}, but solving a nonconvex optimal control problem (OCP) at each time step makes onboard implementation infeasible for resource-limited platforms. Moreover, NMPC performance degrades under model mismatch, which can compromise safety in real-time operation.

Learning-Based (LB) MPC methods mitigate model mismatch by learning residual dynamics from data. Gaussian Process (GP) MPC, in particular, improves prediction accuracy by modeling unmodeled dynamics~\citep{hewing_learning-based_2020, pereida_robust_2021, torrente_data-driven_2021}. Despite demonstrated benefits, GP MPC remains computationally expensive, requiring GP inference along the prediction horizon and solving a nonconvex OCP at each time step. Consequently, existing GP MPC frameworks are executed off-board at rates up to 50~Hz for full-scale multirotors ($>$800~g)~\citep{torrente_data-driven_2021}.

Linear MPC remains the standard for embedded control due to its compatibility with convex solvers and code-generation tools such as OSQP~\citep{banjac_embedded_2017} and CVXGEN~\citep{mattingley_cvxgen_2012}. TinyMPC~\citep{nguyen_tinympc_2024, schoedel_code_2024} introduces an efficient Alternating Direction Method of Multipliers (ADMM)-based solver~\citep{boyd_distributed_2010} optimized for microcontrollers, including onboard deployment on the Crazyflie~2.1. While computationally efficient, these linearized formulations neglect nonlinear effects and model uncertainty, which limits performance and robustness.

To our knowledge, this work presents the first LB MPC deployed \emph{onboard} a tiny multirotor (53~g)—a modified Crazyflie~2.1 platform (Fig.~\ref{fig:teensy}). We introduce \textit{Tiny LB MPC}, a co-designed MPC framework and optimization solver for resource-constrained micro aerial vehicles. Our approach accounts for nonlinear dynamics and model uncertainty while remaining lightweight enough for real-time execution on a microcontroller. Specifically, Tiny LB MPC exploits the multirotor’s differential flatness property~\citep{mellinger_minimum_2011} to separate nonlinear dynamics into equivalent linear models and a nonlinear transformation~\citep{fliess_flatness_1995}, enabling efficient, convex control formulations.

Flatness-Based (FB) MPC~\citep{greeff_flatness-based_2018} leverages this property to decompose control into an outer-loop linear MPC and an inner feedforward linearization block~\citep{hagenmeyer_exact_2003}, yielding convex quadratic programs (QPs). However, most existing implementations depend on general-purpose solvers, limiting achievable control rates and preventing deployment on resource-limited hardware~\citep{akbari_computationally_2024}. Previous work has integrated GPs into MPC to improve model fidelity~\citep{greeff_exploiting_2021, hall_differentially_2023, greeff_learning_2021}, yet these methods neglect how unmodeled dynamics influence input feasibility—effects that introduce nonlinear constraints often omitted in optimization.

The proposed Tiny LB MPC framework integrates an FB MPC formulation with GPs to learn unmodeled dynamics, improving both linearization accuracy and input feasibility handling within the optimization. In parallel, we design a custom ADMM-based solver tailored to the problem structure, enabling real-time computation on embedded hardware. Together, these components achieve high-rate, learning-enhanced predictive control that captures nonlinear effects and uncertainties within a computational budget suitable for microcontrollers.

The contributions of this paper are three-fold:
\begin{itemize}
    \item We present \textit{Tiny LB MPC}, a unified framework combining an MPC formulation that leverages differential flatness and GP-based learning with a solver co-designed for embedded real-time implementation\footnote{An open-source implementation is available at \\https://github.com/Robora-Lab/Tiny-LB-MPC. 
\\A demonstration video is available at\\ https://youtu.be/KmWNd4hREug.}.
    \item We demonstrate the \textit{first onboard deployment of LB MPC} on a 53~g multirotor, achieving 100~Hz control on a Teensy~4.0 microcontroller.
    \item We validate the approach in simulation and hardware experiments, showing up to \textit{43\% improvement in tracking accuracy} over existing embedded MPC baselines under model uncertainty.
\end{itemize}

\section{Problem Statement}
\subsubsection*{Multirotor Dynamics:}
We consider the dynamics of a multirotor described by its position $\mathbf{p}$, velocity $\mathbf{v}$, and rotation matrix $\mathbf{R}$, which represents the orientation of the body frame $B$ with respect to the world frame $W$:
\begin{equation}
    \begin{aligned}
        \dot{\mathbf{p}} &= \mathbf{v},\\
        \dot{\mathbf{v}} &= -g\mathbf{z}_W + \mathbf{c} + f_d(\mathbf{p}, \mathbf{v}, \mathbf{R}),\\
        \dot{\mathbf{R}} &= \mathbf{R}\hat{\pmb{\omega}},
    \end{aligned}
    \label{eq_uav_system}
\end{equation}
where $\hat{\pmb{\omega}}$ is the skew-symmetric matrix of body rates $\pmb{\omega}$, $\mathbf{c} = c\mathbf{z}_B$ is the mass-normalized thrust vector in the world frame, and $f_d(\cdot)$ captures unknown dynamics, including internal modeling errors (e.g., additional mass, thrust--PWM nonlinearities) and external disturbances (e.g., drag, ground effect). The system state and input are defined as $\mathbf{x} = [\mathbf{p}, \mathbf{v}, \mathbf{R}]^\top$ and $\mathbf{u} = [c, \pmb{\omega}]^\top$, respectively.

Assuming $f_d(\mathbf{x}) = 0$, the model~(\ref{eq_uav_system}) is \textit{differentially flat} with flat output $\mathbf{y} = [\mathbf{p}, \psi]^\top$, where $\psi$ denotes the yaw angle~\citep{mellinger_minimum_2011}. This property allows the nonlinear dynamics to be transformed via feedback or feedforward linearization~\citep{greeff_flatness-based_2018, greeff_exploiting_2021} into an equivalent linear representation with flat state $\mathbf{z} = [\mathbf{p}, \mathbf{v}, \mathbf{a}, \psi]^\top \in \mathbb{R}^{10}$ and flat input $\mathbf{v} = [\mathbf{j}, \dot{\psi}]^\top$, where $\mathbf{j}$ is the jerk. After discretization, the flat-space dynamics become:
\begin{equation}
    \mathbf{z}_{i+1} = \mathbf{A}\mathbf{z}_i + \mathbf{B}\mathbf{v}_i,
    \label{eq_lin_flat}
\end{equation}
where $i$ denotes the discrete time step.

\begin{assumption}
\label{assumption1}
The system~(\ref{eq_uav_system}) retains the differential flatness property despite the presence of the disturbance term $f_d(\mathbf{x})$.
\end{assumption}

Assumption~\ref{assumption1} implies the existence of smooth, invertible mappings---i.e., \textit{diffeomorphisms}---between the original system state $\mathbf{x}$ and the flat state $\mathbf{z}$, as well as between the inputs $\mathbf{u}$ and $\mathbf{v}$~\citep{fliess_flatness_1995}. Consequently, the unknown dynamics can be expressed as a function of the flat state, $f_d(\mathbf{x}) = d(\mathbf{z})$, where $d(\mathbf{z})$ represents unmodeled dynamics learned using a Gaussian Process (GP). As discussed in Sec.~\ref{sec_methodology}, the learned model $d(\mathbf{z})$ influences both feedforward linearization (i.e., the mapping from $\mathbf{z}$ and $\mathbf{v}$ to $\mathbf{u}$) and the enforcement of input feasibility (e.g., thrust vector constraints) within the MPC optimization.

\subsubsection*{Optimal Control Problem:}
Building on the flatness-based representation and the learned uncertainty model $d(\mathbf{z})$, we formulate the predictive control problem to generate dynamically feasible and safe trajectories under uncertainty while maintaining computational efficiency for embedded deployment.

We define the Learning-Based MPC problem as:
\begin{equation}
\begin{aligned}
\min_{\mathbf{z}_{1:N}, \mathbf{v}_{0:N-1}} \quad & J(\mathbf{z}_{1:N}, \mathbf{v}_{0:N-1}) \\
\text{s.t.} \quad &
\mathbf{z}_{k+1} = \mathbf{A}\mathbf{z}_k + \mathbf{B}\mathbf{v}_k, \quad \forall k \in \mathcal{K},\\
& \mathbf{z}_k \in \mathcal{Z}, \quad \forall k \in \mathcal{K},
\end{aligned}
\label{eq_ocp}
\end{equation}
where $\mathcal{K} = \{0, 1, \ldots, N-1\}$ and $N$ is the prediction horizon. The set $\mathcal{Z}$ encodes the physical limits of the multirotor, including maximum collective thrust and allowable attitude angles. To ensure safety under model uncertainty, these constraints are expressed probabilistically on the mass-normalized thrust vector $\mathbf{c}_k = [c^x_k, c^y_k, c^z_k]^\top$:
\begin{equation}
    \text{Pr}(\mathbf{c}_k \in S_{\text{ball}}) \geq p_b, \quad \forall k \in \mathcal{K}
     \label{eq_ball}
\end{equation}
\begin{equation}
    \text{Pr}(\mathbf{c}_k \in S_{\text{cone}}) \geq p_c, \quad \forall k \in \mathcal{K}
    \label{eq_cone}
\end{equation}
where $p_b$ and $p_c$ are user-defined confidence levels. The feasible thrust regions are defined as:
\begin{equation*}
    S_{\text{ball}} = \{\mathbf{c}_k \mid \lVert\mathbf{c}_k\rVert_2 \leq c_{\text{max}} \},
\end{equation*}
\begin{equation*}
    S_{\text{cone}} = \{\mathbf{c}_k \mid \lVert[c^x_{k}, c^y_{k}]^\intercal\rVert_2 \leq c^z_{k} \tan{\theta_{\text{max}}}\},
\end{equation*}
where $c_{\text{max}}$ is the maximum mass-normalized collective thrust, and $\theta_{\text{max}}$ is the maximum pitch or roll angle allowed for safe operation.

The objective function penalizes tracking error and control effort using a standard quadratic cost:
\begin{equation}
\begin{aligned}
J(\cdot) &= \frac{1}{2}\mathbf{z}_N^\top \mathbf{Q}_f \mathbf{z}_N + \mathbf{q}_f^\top \mathbf{z}_N \\
&\quad + \sum_{k=0}^{N-1} 
\left[
\frac{1}{2}\mathbf{z}_k^\top \mathbf{Q}_k \mathbf{z}_k + \mathbf{q}_k^\top \mathbf{z}_k
+ \frac{1}{2}\mathbf{v}_k^\top \mathbf{R}\mathbf{v}_k + \mathbf{r}_k^\top \mathbf{v}_k
\right],
\end{aligned}
\label{eq_quadratic_cost}
\end{equation}
where $\mathbf{Q}_k \succeq 0$, $\mathbf{Q}_f \succeq 0$, and $\mathbf{R} \succ 0$ are symmetric weighting matrices, and $\mathbf{q}_k$, $\mathbf{r}_k$ are linear cost terms that may vary to follow a reference trajectory.

\section{Background}

\label{sec_background}
To provide context for our
LB MPC
formulation, we review
two classical control
methods that underpin
our approach. We first
outline the unconstrained
Linear Quadratic
Regulator (LQR)
formulation, which admits
a closed-form analytical
solution, and then discuss
how convex constraints
can be incorporated
efficiently using ADMM.

\subsubsection*{Linear Quadratic Regulator:}
\label{sec_lqr}
To gain insight into the
structure of the OCP (\ref{eq_ocp}), we first
consider the unconstrained case where state constraints are neglected reducing the
formulation to a standard finite-horizon LQR. Neglecting state constraints $\mathbf{z}_{k} \in \mathcal{Z}$ in (\ref{eq_ocp}), the OCP can be treated as a finite-horizon discrete-time LQR and has a closed-form solution in the form of a linear feedback controller \citep{lewis_optimal_2012}:
\vspace{-0.1cm}
\begin{equation}
    \mathbf{v}_k = - \mathbf{K}_k \mathbf{z}_k - \mathbf{d}_k,
    \label{eq_control_seq}
\end{equation}
where the gain matrix $\mathbf{K}_k$ and feedforward term $\mathbf{d}_k$ can be computed by setting $\mathbf{P}_N = \mathbf{Q}_f$, $\mathbf{p}_N = \mathbf{q}_f$, and  solving the discrete Riccati equation (\ref{eq_riccati}) backward in time \citep{lewis_optimal_2012}:

\vspace{-0.5cm}
\begin{align}
\label{eq_riccati}
\mathbf{K}_k &= (\mathbf{R} + \mathbf{B}^\intercal \mathbf{P}_{k+1} \mathbf{B})^{-1}(\mathbf{B}^\intercal \mathbf{P}_{k+1} \mathbf{A}), \nonumber\\
\mathbf{d}_k &= (\mathbf{R} + \mathbf{B}^\intercal \mathbf{P}_{k+1} \mathbf{B})^{-1}(\mathbf{B}^\intercal\mathbf{p}_{k+1} + \mathbf{r}_k), \\ 
\mathbf{P}_k &= \mathbf{Q}_k + \mathbf{K}_{k}^{\intercal} \mathbf{R} \mathbf{K}_{k} 
+ (\mathbf{A}-\mathbf{B} \mathbf{K}_k)^\intercal \mathbf{P}_{k+1}(\mathbf{A}- \mathbf{B}\mathbf{K}_k), \nonumber\\
\mathbf{p}_k &= \mathbf{q}_k + (\mathbf{A}-\mathbf{B} \mathbf{K}_k)^\intercal 
(\mathbf{p}_{k+1} - \mathbf{P}_{k+1}\mathbf{B} \mathbf{d}_k) \nonumber\\
&\quad + \mathbf{K}_k^\intercal(\mathbf{R}\mathbf{d}_k - \mathbf{r}_k). \nonumber
\end{align}
\subsubsection*{Alternating Direction Method of Multipliers:} When state constraints are reintroduced, the OCP no longer admits a closed-form solution, motivating the use of numerical optimization methods.
If the state constraints $\mathbf{z}_{k} \in \mathcal{Z}$ in (\ref{eq_ocp}) are convex, then a popular approach to solving the OCP is to use ADMM. ADMM can be used to solve general convex problems of the following form \citep{boyd_distributed_2010}:
\begin{equation}
\begin{aligned}
	\min_{\mathbf{t}} & \quad g(\mathbf{t}) \\
\textrm{s.t.} \quad & \mathbf{t} \in \mathcal{C},
  \label{eq_convex_ocp}
\end{aligned}
\end{equation}
with cost $g$ and $\mathcal{C}$ convex. The indicator function for set $\mathcal{C}$ is defined:
\begin{equation}
    I_{\mathcal{C}}(\mathbf{s}) = \begin{cases}
0 &\text{$\mathbf{s} \in \mathcal{C}$}\\
\infty &\text{otherwise.}
\end{cases}
\label{eq_indicator}
\end{equation}
We can now form the following equivalent problem by introducing the slack variable $\mathbf{s}$:
\begin{equation}
\begin{aligned}
\min_{\mathbf{t}, \mathbf{s}} & \quad g(\mathbf{t}) + I_{\mathcal{C}}(\mathbf{s}) \\
\textrm{s.t.} \quad & \mathbf{t} = \mathbf{s}.
\label{eq_admm_p}
\end{aligned}
\end{equation}
The augmented Lagrangian, with Lagrange multiplier $\pmb{\lambda}$ and penalty parameter $\rho$, of the transformed problem (\ref{eq_admm_p}) is:
\vspace{-0.1cm}
\begin{equation}
    \mathcal{L}_{A}(\mathbf{t}, \mathbf{s}, \pmb{\lambda}) = g(\mathbf{t}) + I_{\mathcal{C}}(\mathbf{s}) + \pmb{\lambda}^\intercal(\mathbf{t} - \mathbf{s}) + \frac{\rho}{2} \lVert\mathbf{t} - \mathbf{s}\rVert_2^2.
\label{eq_augmented_lagrangian}
\end{equation}
ADMM alternates minimization over primal $\mathbf{t}$, slack $\mathbf{s}$ and dual variables (Lagrange multipliers):
\begin{equation}
    \text{primal update: } \mathbf{t}^+ = \underset{\mathbf{t}}{\operatorname{argmin}} \quad \mathcal{L}_{A}(\mathbf{t}, \mathbf{s}^+, \pmb{\lambda}^+),
    \label{eq_primalupdate}
\end{equation}
\vspace{-0.3cm}
\begin{equation}
    \text{slack update: } \mathbf{s}^+ = \underset{\mathbf{s}}{\operatorname{argmin}} \quad \mathcal{L}_{A}(\mathbf{t}^+, \mathbf{s}, \pmb{\lambda}^+),
    \label{eq_slackupdate}
\end{equation}
\vspace{-0.3cm}
\begin{equation}
    \text{dual update: } \pmb{\lambda}^+ = \pmb{\lambda} + \rho (\mathbf{t}^+ - \mathbf{s}^+),
    \label{eq_dualupdate1}
\end{equation}
where (\ref{eq_dualupdate1}) is a gradient-ascent update on the Lagrange multiplier. These steps (\ref{eq_primalupdate})-(\ref{eq_dualupdate1}) can be repeated until the desired level of convergence is reached.

\subsubsection*{Second-Order Conic Projection:}

To further improve the computational
efficiency of constrained optimization
within ADMM, we recall a key operation
required when handling convex constraints
expressed as second-order cones (SOCs).
In such cases, the projection step—central
to the slack variable update—can be
evaluated analytically rather than through
iterative solvers, greatly accelerating
convergence in real-time MPC
applications.
The projection of $(s, \mathbf{s}) \in \mathbb{R} \times \mathbb{R}^n$ onto the SOC:
\begin{equation}
    \mathcal{C} = \{(t, \mathbf{t})\in \mathbb{R} \times \mathbb{R}^n : \lVert\mathbf{t} \rVert_2 \leq t \},
    \label{eq_soc_def}
\end{equation}
has a closed-form solution:
\begin{equation}
    \text{proj}_{\mathcal{C}}(s, \mathbf{s}) = \begin{cases}
0, & \lVert\mathbf{s} \rVert_2 \leq -s \\
(s, \mathbf{s}) & \lVert\mathbf{s} \rVert_2 \leq s \\ 
\frac{s + \lVert\mathbf{s} \rVert_2}{2 \lVert\mathbf{s} \rVert_2}(\lVert\mathbf{s} \rVert_2, \mathbf{s}) & \lVert\mathbf{s} \rVert_2 > |s|.
\end{cases}
\label{eq_soc_closed}
\end{equation}
As demonstrated in \citep{schoedel_code_2024}, if convex constraints $\mathcal{C}$ in (\ref{eq_indicator}) are described by  SOCs  of the form (\ref{eq_soc_def}) then the slack update of ADMM (\ref{eq_slackupdate}) can be efficiently computed using the closed form solution (\ref{eq_soc_closed}).

\subsubsection*{Gaussian Processes:}

To model the unknown component of the system dynamics in
(\ref{eq_uav_system}) and account for uncertainty in prediction and control, we
employ GP regression, a probabilistic and
nonparametric learning framework. This approach provides
both a mean and variance estimate for the learned
disturbance term, enabling the probabilistic constraints in
(\ref{eq_ball})–(\ref{eq_cone}) to be reformulated as convex SOC
constraints on the flat state, thereby maintaining tractability
within the MPC formulation. GP regression is a nonparametric method for approximating a nonlinear function \(d(\mathbf{z}): \mathbb{R}^{\dim(\mathbf{z})} \rightarrow \mathbb{R}\), where each function value \(d(\mathbf{z})\) is a random variable and any finite collection follows a joint Gaussian distribution. Following the notation in  Sec. \ref{sec_methodology} , we denote the input as \(\mathbf{z}\) and the output as \(d(\mathbf{z})\). A GP is defined by a prior mean—typically zero—and a kernel function \(k(\cdot, \cdot)\) that encodes similarity between inputs. A common choice is the squared-exponential (SE) kernel:
\[
k(\mathbf{z}_l,\mathbf{z}_m) = \sigma_{\eta}^2\exp{\left(-\tfrac{1}{2}(\mathbf{z}_l-\mathbf{z}_m)^\intercal\mathbf{M}^{-2}(\mathbf{z}_l -\mathbf{z}_m)\right)} + \delta_{lm}\sigma_{\omega}^2,
\]
where \(\sigma^2_{\eta}\) is the prior variance, \(\sigma^2_{\omega}\) is the measurement noise (with \(\delta_{lm} = 1\) if \(l = m\), 0 otherwise), and \(\mathbf{M}\) contains the length scales that govern how quickly \(d(\mathbf{z})\) varies with \(\mathbf{z}\) \citep{rasmussen_gaussian_2008}.
\section{Methodology}
\label{sec_methodology}
\begin{figure}[t]
\centering
\includegraphics[width=1\linewidth, trim={2cm 2cm 2.5cm 2cm}]{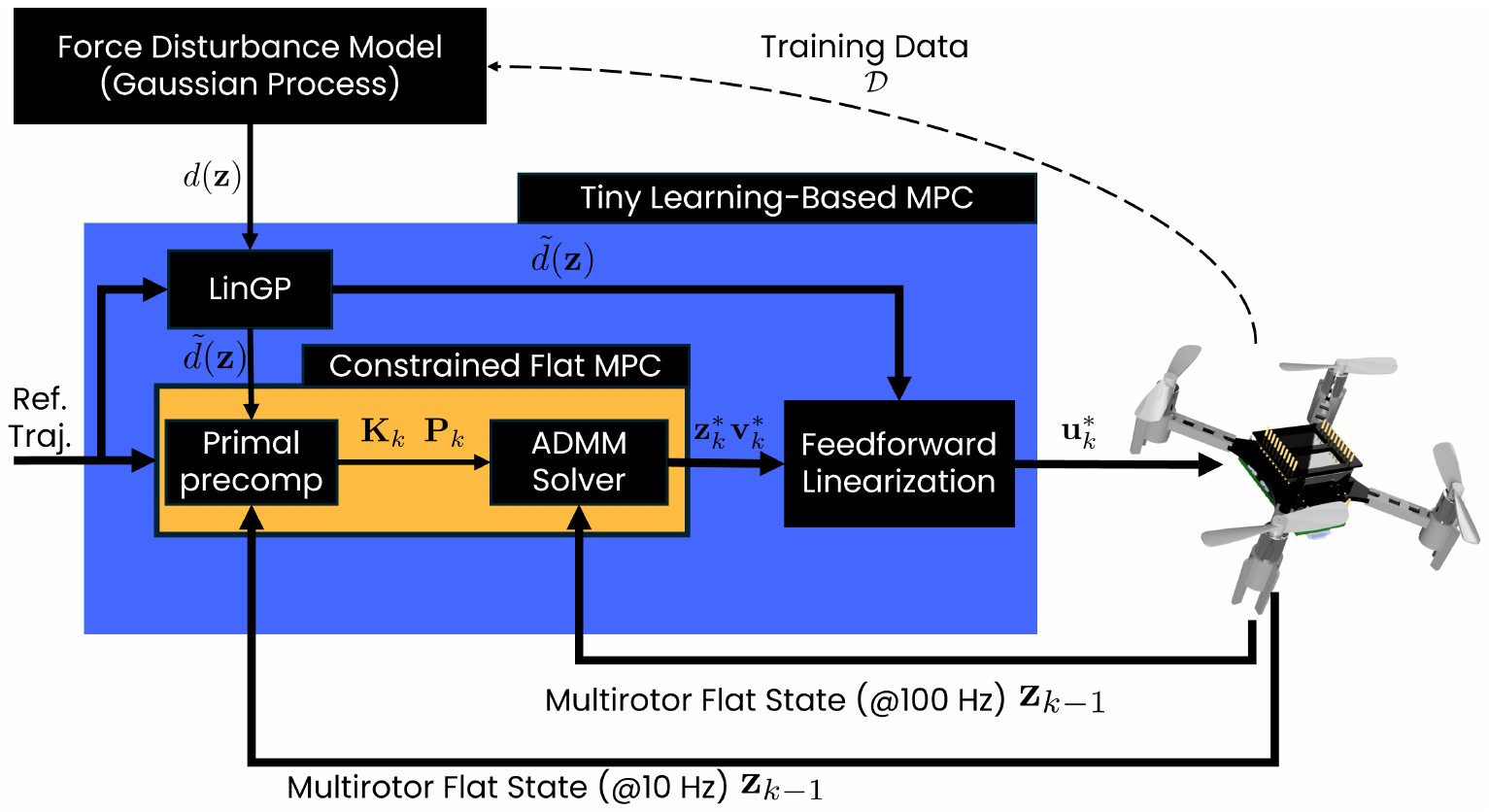}
  \vspace{-1.2cm}
    \caption{{Block diagram of the proposed Tiny~LB~MPC framework. 
The controller integrates the multirotor’s differential flatness, 
a Linearized Gaussian Process (LinGP) model of residual dynamics, 
and a linear MPC with second-order cone (SOC) constraints to compute 
the control input $\mathbf{u}$ at 100~Hz on the Teensy~4.0 microcontroller. }}
\label{fig:system}
\end{figure}

To compensate for both internal model errors and external disturbances, we learn the unknown component of the dynamics, \( f_d(\mathbf{x}) = d(\mathbf{z}) \), as a GP function of the flat state \( \mathbf{z} \). Specifically, we decompose \( d(\mathbf{z}) = [d_x(\mathbf{z}), d_y(\mathbf{z}), d_z(\mathbf{z})]^\top \), where each component represents the learned disturbance in the respective coordinate direction. Our \textit{Tiny Learning-Based MPC} framework (Fig.~\ref{fig:system}) integrates four key components that together enable real-time control at 100~Hz on a resource-constrained microcontroller. The  four key components are: 
(i) learning the residual dynamics as a Linearized Gaussian Process (LinGP), 
(ii) reformulating probabilistic constraints as deterministic SOCs, 
(iii) implementing a custom ADMM-based solver optimized for embedded execution, and 
(iv) feedforward linearization for thrust command computation.

\subsection{Learning the Disturbance using LinGP}
To maintain computational efficiency while leveraging learned dynamics, we introduce a LinGP model by linearizing the full GP \( d(\mathbf{z}) \) about a reference flat state \(\mathbf{z}_r \). The resulting LinGP retains the probabilistic nature of the GP but exhibits a simplified structure: its posterior mean is affine and its covariance is quadratic in \( \mathbf{z} \). This structure preserves uncertainty information while significantly reducing inference cost, making it suitable for onboard deployment.

The LinGP posterior, conditioned on trajectory data \( \mathcal{D} \), is expressed as:
\begin{equation}
    \tilde{d}(\mathbf{z}) \mid \mathcal{D} \sim \mathcal{N}(\tilde{\mu}(\mathbf{z}), \tilde{\Sigma}(\mathbf{z})),
    \label{eq:lingp_def}
\end{equation}
where the posterior mean is
\begin{equation}
    \tilde{\mu}(\mathbf{z}) = \bar{\mu}^\top(\mathbf{z}_r)\,\bar{\mathbf{z}}, \quad \text{with} \quad \bar{\mathbf{z}} = [1,\, \mathbf{z} - \mathbf{z}_r]^\top,
    \label{eq:lingp_mean}
\end{equation}
and the covariance is diagonal with entries quadratic in \( \mathbf{z} \):
\begin{equation}
    \tilde{\Sigma}(\mathbf{z}) = \mathrm{diag}\big(
    \bar{\mathbf{z}}^\top \bar{V}_x(\mathbf{z}_r)\bar{\mathbf{z}},\,
    \bar{\mathbf{z}}^\top \bar{V}_y(\mathbf{z}_r)\bar{\mathbf{z}},\,
    \bar{\mathbf{z}}^\top \bar{V}_z(\mathbf{z}_r)\bar{\mathbf{z}}
    \big).
    \label{eq:lingp_cov}
\end{equation}
Here, the matrices \( \bar{V}_x(\mathbf{z}_r), \bar{V}_y(\mathbf{z}_r), \bar{V}_z(\mathbf{z}_r) \) and coefficients \( \bar{\mu}(\mathbf{z}_r) \) are obtained from GP inference at the reference state \( \mathbf{z}_r \), see \citep{akbari_computationally_2024}. This linearized representation preserves the probabilistic structure of the GP while maintaining computational tractability, enabling real-time inference on embedded hardware.

\subsection{Thrust Constraints as Second-Order Cones
}

\label{sec_b}

Using the LinGP model, we reformulate the probabilistic thrust constraints in \eqref{eq_ball}--\eqref{eq_cone} as deterministic SOC constraints on the flat state \( \mathbf{z} \). This transformation converts the stochastic OCP \eqref{eq_ocp} into a tractable convex program while maintaining safety guarantees. The GP-based tightening ensures that each constraint incorporates the predicted uncertainty in \( d(\mathbf{z}) \), with tightening factors determined by quantiles of the chi-squared and Gaussian distributions~\citep{hewing_cautious_2020}. 

The learned LinGP model allows us to express the mass-normalized thrust vector as:
\begin{equation}
    \mathbf{c} = \mathbf{a} + g \mathbf{z}_W - \tilde{d}(\mathbf{z}),
    \label{eq:thrust_vector_gp}
\end{equation}
where \( \tilde{d}(\mathbf{z}) \) is distributed according to the posterior in~\eqref{eq:lingp_def}. Each component of \( \mathbf{c} \) therefore has a posterior mean that is affine in \( \mathbf{z} \) and a covariance that is quadratic in \( \mathbf{z} \).

\subsubsection{Thrust Magnitude Constraints:}
We first consider the probabilistic thrust-magnitude constraint in~\eqref{eq_ball}, which ensures that the total thrust magnitude does not exceed the physical limit of the rotors. Using the LinGP approximation, this constraint can be rewritten for each time step \( k \in \mathcal{K} \) as a deterministic constraint tightened by the GP variance:
\begin{equation}
    \| \mathbf{a}_k + g \mathbf{z}_W - \bar{\mu}^\top(\mathbf{z}_{r_k}) \bar{\mathbf{z}}_k \|_2 
    \le c_{\max} - \beta \sqrt{\lambda_{\max}(\tilde{\Sigma}(\mathbf{z}_k))},
    \label{eq:soc_ball_tightened}
\end{equation}
where \( \beta = \sqrt{\chi^2_3(p_b)} \) is the quantile of the chi-squared distribution with three degrees of freedom corresponding to the desired confidence level \( p_b \), and \( \lambda_{\max}(\cdot) \) denotes the maximum eigenvalue of the covariance matrix. Because the covariance \( \tilde{\Sigma}(\mathbf{z}_k) \) is diagonal, this maximum eigenvalue lies on the diagonal, and the constraint can be equivalently decomposed into three scalar constraints along the \( x \), \( y \), and \( z \) axes.

To express~\eqref{eq:soc_ball_tightened} in a standard SOC form, we introduce six auxiliary variables \( \gamma_{1k}, \ldots, \gamma_{6k} \) such that, for \( j \in \mathbb{Z} \cap [1,3] \),
\begin{equation}
    \gamma_{jk} = c_{\max} - \gamma_{(3+j)k}.
    \label{eq:gamma_relation_ball}
\end{equation}
The deterministic tightening of the thrust-magnitude constraint can then be expressed as:
\begin{align}
    \| \mathbf{a}_k + g \mathbf{z}_W - \bar{\mu}^\top(\mathbf{z}_{r_k}) \bar{\mathbf{z}}_k \|_2 &\le \gamma_{jk}, 
    && j \in [1,3], \label{eq:soc_thrust_magnitude} \\[2pt]
    \| \bar{L}_w^\top(\mathbf{z}_{r_k}) \bar{\mathbf{z}}_k \|_2 &\le \gamma_{(3+j)k}, 
    && j \in [1,3], \label{eq:soc_variance}
\end{align}
where \( w = x, y, z \) corresponds to the respective component of the thrust vector, and \( \bar{L}_w(\mathbf{z}_{r_k}) \) is the Cholesky factor of the variance matrix \( \bar{V}_w(\mathbf{z}_{r_k}) \), i.e.,
\[
    \sqrt{\bar{\mathbf{z}}_k^\top \bar{V}_w(\mathbf{z}_{r_k}) \bar{\mathbf{z}}_k} = 
    \| \bar{L}_w^\top(\mathbf{z}_{r_k}) \bar{\mathbf{z}}_k \|_2.
\]
Equations~\eqref{eq:soc_thrust_magnitude}--\eqref{eq:soc_variance}, together with the equality relation in~\eqref{eq:gamma_relation_ball}, represent six SOC constraints and three linear equality constraints per time step \( k \).

\subsubsection{Thrust Angle Constraints:}
We now consider the probabilistic thrust-angle constraint in~\eqref{eq_cone}, which bounds the lateral thrust components relative to the vertical component to ensure safe pitch and roll limits. Using the LinGP model, this constraint can be similarly reformulated as a set of deterministic SOC and linear inequality constraints.

For each time step \( k \), we define auxiliary variables \( \gamma_{7k}, \gamma_{8k}, \gamma_{9k} \) and a selection vector \( h = [1, 1, 0]^\top \) to isolate the horizontal thrust components. The SOC constraints are then given by:
\begin{align}
    \| h^\top (\mathbf{a}_k - \bar{\mu}^\top(\mathbf{z}_{r_k}) \bar{\mathbf{z}}_k) \|_2 
    &\le \gamma_{jk}, && j \in [8,9], \label{eq:soc_angle_1}
\end{align}
where the auxiliary variables satisfy the following linear relations:
\begin{equation}
    \gamma_{jk} = \gamma_{7k} - \beta \gamma_{(j-4)k}, \quad j \in [8,9].
    \label{eq:gamma_relation_angle}
\end{equation}
Finally, we impose a linear inequality constraint to ensure that the vertical thrust component does not exceed the limit imposed by the maximum tilt angle \( \theta_{\max} \):
\begin{equation}
    \gamma_{6k} \le 
    \alpha \!\left(
        \bar{\mu}_z^\top(\mathbf{z}_{r_k}) \bar{\mathbf{z}}_k 
        - \frac{1}{\tan(\theta_{\max})} \gamma_{7k}
    \right),
    \label{eq:linear_angle_constraint}
\end{equation}
where \( \alpha = 1 / \Phi^{-1}(p_c) \) and \( \Phi^{-1} \) is the inverse of the standard Gaussian cumulative distribution function (CDF) corresponding to the probability threshold \( p_c \).

\subsubsection{Summary of Constraints:}
Equations~\eqref{eq:soc_thrust_magnitude}--\eqref{eq:linear_angle_constraint} define the tightened deterministic reformulation of the probabilistic constraints in \eqref{eq_ball}--\eqref{eq_cone}. At each time step \( k \), these include: eight SOC constraints on the thrust magnitude and angle limit with the GP variance, five linear equalities relating auxiliary variables for tightening, and one linear inequality. Together, these constraints ensure that the feasible thrust region accounts for both the learned uncertainty and the physical actuation limits of the multirotor, resulting in a safe and convex formulation suitable for embedded optimization.

\subsection{Tiny Learning-Based MPC Solver}
\label{sect_solver}
Including the LinGP uncertainty model and tightened SOC constraints converts the MPC into a convex Second-Order Cone Program (SOCP). Executing such problems at $100$~Hz on the 600~MHz Cortex-M7 of the Teensy~4.0 requires a solver that is both memory-efficient and numerically predictable. General-purpose solvers such as OSQP exceed the available resources; therefore, we design a custom ADMM solver that exploits the block structure of our MPC formulation.

We therefore employ the ADMM framework, which decomposes the global optimization into smaller subproblems that can be solved in closed form. The specific problem structure allows each step of ADMM to be expressed using simple matrix–vector operations, making it suitable for embedded real-time implementation.

The solver’s key idea is to decouple the optimization of the state trajectory 
$\mathbf{z}_{1:N}$ and inputs $\mathbf{v}_{0:N-1}$ 
from the dummy variables 
$\pmb{\gamma}_{k} = [\gamma_{1k}, \ldots, \gamma_{9k}]^\top$.
Since there are no equality constraints linking these two groups of variables,
ADMM can be applied to compute 
two independent primal updates—one for the state–input block and one for the dummy variables.

\subsubsection*{Problem Structure:}
The optimization problem takes the form:
\begin{equation}
\min_{\mathbf{z}_{1:N}, \mathbf{v}_{0:N-1}, \pmb{\gamma}_{0:N}}
J(\mathbf{z}_{0:N}, \mathbf{v}_{0:N-1})
\label{eq_tinylb_cost}
\end{equation}
subject to linear dynamics and the constraint sets:
\begin{align}
\text{C1: } & \|\mathbf{F}_{jk}\mathbf{z}_k + \mathbf{g}_{jk}\|_2 \le \gamma_{jk}, \nonumber\\
\text{C}_\gamma: & \;\mathbf{C}\pmb{\gamma}_k = \mathbf{d}, \nonumber\\
\text{C2: } & \;\mathbf{T}\pmb{\gamma}_k \le \mathbf{S}_k\mathbf{z}_k + \mathbf{w}_k,
\label{eq_tinylb_form}
\end{align}
where $J(\cdot)$ is the quadratic cost from~(\ref{eq_quadratic_cost}). The SOCs (C1) encode thrust-magnitude and attitude constraints informed by LinGP, i.e., Equations~\eqref{eq:soc_thrust_magnitude}--\eqref{eq:soc_angle_1}, the linear equalities (C$_\gamma$) couple auxiliary variables $\gamma_{1\ldots9}$, i.e., Equations~\eqref{eq:gamma_relation_ball} and~\eqref{eq:gamma_relation_angle}, and the linear inequality (C2) represents the maximum tilt angle, i.e., Equation~\eqref{eq:linear_angle_constraint}. Crucially, no equality directly couples the state--input variables $(\mathbf{z}_{1:N}, \mathbf{v}_{0:N-1})$ and the dummy variables $\pmb{\gamma}_k$. This separation enables ADMM to perform two independent primal updates---one for the trajectory and one for the auxiliary variables---reducing memory access and iteration time.

\subsubsection*{ADMM Reformulation:}
Following the standard ADMM procedure, we introduce slack variables $\pmb{\zeta}_{jk}, \tau_{jk}$ for C1 and $\pmb{\rho}_k, \pmb{\sigma}_k$ for C2, yielding:
\begin{align}
\text{C3: } & \mathbf{F}_{jk}\mathbf{z}_k + \mathbf{g}_{jk} = \pmb{\zeta}_{jk}, &
\text{C4: } & \;\gamma_{jk} = \tau_{jk}, \nonumber\\
\text{C5: } & \mathbf{C}\pmb{\gamma}_k = \mathbf{d}, &
\text{C6: } & \;\mathbf{T}\pmb{\gamma}_k = \pmb{\rho}_k, \nonumber\\
\text{C7: } & \mathbf{S}_k\mathbf{z}_k + \mathbf{w}_k = \pmb{\sigma}_k.
\label{eq_admm_reform}
\end{align}
The cost is augmented with indicator functions $I_{\mathcal{C}_1}(\pmb{\zeta}, \tau)$ and $I_{\mathcal{C}_2}(\pmb{\rho}, \pmb{\sigma})$ enforcing cone and half-space feasibility. The resulting augmented Lagrangian follows the standard form of~(\ref{eq_augmented_lagrangian}). Each ADMM iteration alternates between:

\begin{enumerate}[label=(\roman*)]
    \item \textbf{State--input primal update:} $(\mathbf{z}_{1:N}, \mathbf{v}_{0:N-1})$ are updated via linear equations resembling a finite-horizon LQR backward/forward pass (Sec. \ref{sec_lqr}).
    \item \textbf{Dummy-variable primal update:} $\pmb{\gamma}_k$ is updated in closed form using the affine constraint (C5).
    \item \textbf{Slack updates:} $(\pmb{\zeta}, \tau)$ are projected onto the SOC using the closed-form rule~(\ref{eq_soc_closed}), while $(\pmb{\rho}, \pmb{\sigma})$ are projected onto the half-space via simple clipping.
    \item \textbf{Dual updates:} Lagrange multipliers for C3--C7 are updated via gradient ascent following~(\ref{eq_dualupdate1}).
\end{enumerate}

\subsubsection*{Pre-Computation and Embedded Execution:}
To accelerate the state--input primal update, a \textit{low-rate pre-computation loop} ($\approx 10$~Hz) computes time-varying feedback gains $\mathbf{K}_k$ and cost-to-go matrices $\mathbf{P}_k$ by solving the discrete Riccati equation~(\ref{eq_riccati}) with modified weights. For compactness, we define stacked constraint matrices and vectors as:
\[
\mathbf{F}_k = \mathrm{blkdiag}(\mathbf{F}_{1k}, \dots, \mathbf{F}_{Jk}),
\]
and $\mathbf{g}_k = [\mathbf{g}_{1k}^\top, \dots, \mathbf{g}_{Jk}^\top]^\top, \quad
\pmb{\zeta}_k = [\pmb{\zeta}_{1k}^\top, \dots, \pmb{\zeta}_{Jk}^\top]^\top.
$
Using these stacked variables, the modified cost weights used in the Riccati recursion are computed as:
\begin{equation}
\begin{aligned}
\tilde{\mathbf{Q}}_k &= \mathbf{Q}_k + \rho(\mathbf{F}_k^\top\mathbf{F}_k + \mathbf{S}_k^\top\mathbf{S}_k),\\
\tilde{\mathbf{q}}_k &= \mathbf{q}_k + \mathbf{F}_k^\top(\pmb{\lambda}_k + \rho(\mathbf{g}_k - \pmb{\zeta}_k))
+ \mathbf{S}_k^\top(\pmb{\Lambda}_k + \rho(\mathbf{w}_k - \pmb{\sigma}_k)),
\label{eq_modified_weights}
\end{aligned}
\end{equation}
where $\pmb{\lambda}_{k}$ and $\pmb{\Lambda}_k$ are the Lagrange multipliers for the equality constraints (C3) and (C7) respectively.
This pre-computation incorporates the GP-inferred matrices $\mathbf{F}_{jk}, \mathbf{g}_{jk}, \mathbf{S}_k, \mathbf{w}_k$ to adapt $\mathbf{K}_k$ and $\mathbf{P}_k$ to the predicted uncertainty $\tilde{d}(\mathbf{z}_k)$. The \textit{high-rate solver loop} ($\approx 100$~Hz) then uses the feedback gain $\mathbf{K}_k$ and cost-to-go matrices $\mathbf{P}_k$ and executes 3--5 ADMM iterations comprising the Riccati backward pass, forward rollout, SOC projection, and dual updates---all using fixed-size vector--matrix operations and no dynamic memory allocation.

Through this solver--model co-design, the Tiny LB-MPC executes complete SOCP updates onboard within 8~ms per control cycle, enabling real-time, learning-enhanced predictive control on a $53$~g Crazyflie platform.

\subsection{Feedforward Linearization}

The feedforward block, see Fig. \ref{fig:system},  bridges the high-level flat state trajectory and the low-level thrust commands by compensating for the learned uncertainty. Starting from the nominal flat state acceleration $\mathbf{a}$ generated by the MPC, we refine the mapping to the commanded thrust input $\mathbf{c}$ by incorporating LinGP’s predicted mean $\tilde{\mu}(\mathbf{z})$ in (\ref{eq:lingp_mean}), yielding:
\begin{equation}
\mathbf{c} = \mathbf{a} + g\mathbf{z}_W - \tilde{\mu}(\mathbf{z}).
\label{eq_feedforward}
\end{equation}
While the feedforward linearization primarily leverages the mean prediction for compensation, the corresponding GP uncertainty is still accounted for through the thrust and attitude feasibility constraints in our MPC formulation (Sec.~\ref{sec_b}).  To compute the control input $\mathbf{u}$, the SOCP in \eqref{eq_tinylb_cost}-\eqref{eq_tinylb_form} is solved using our custom ADMM solver (Sec.~\ref{sect_solver}), after which the final thrust command is updated via (\ref{eq_feedforward}) following the flatness-based linearization approach of \citep{greeff_flatness-based_2018}. This feedforward stage closes the loop between flat-state optimization and actuator commands, ensuring that learned dynamics are consistently compensated in both prediction and control execution.

\begin{figure}[t]
     \centering    
    \includegraphics[width=0.8\linewidth]{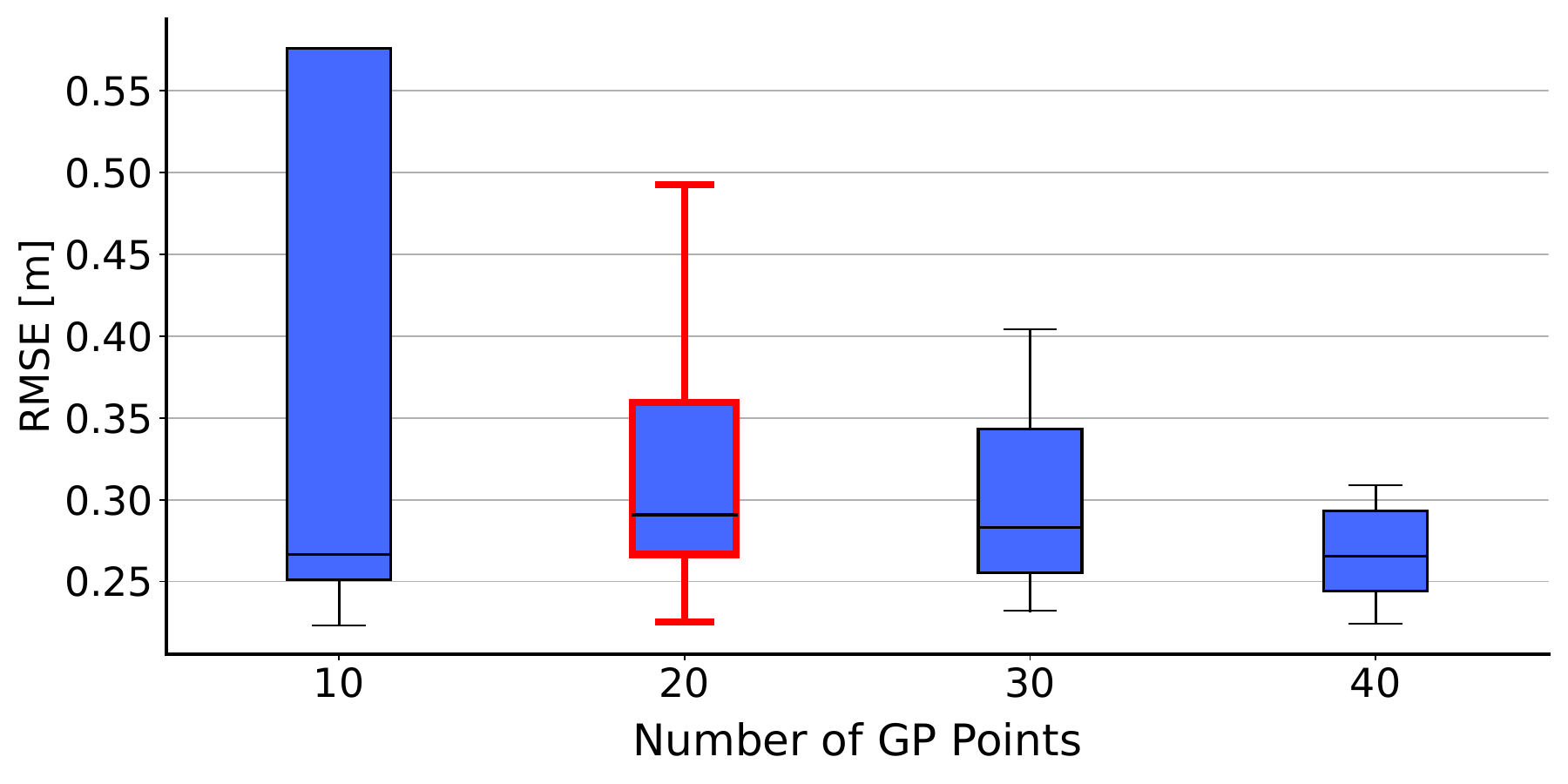}
    \vspace{-0.4cm}
     \caption{\textit{Simulation 1 —} Effect of GP training data size on RMSE for a 3~rad/s circular trajectory over 20 trials. The solver runs 5 ADMM iterations; 20 data points achieve low RMSE with minimal variance.}
     \label{fig:rmse_boxplot}
 \end{figure}
 \begin{figure}
    \centering
    \includegraphics[width=1\linewidth]{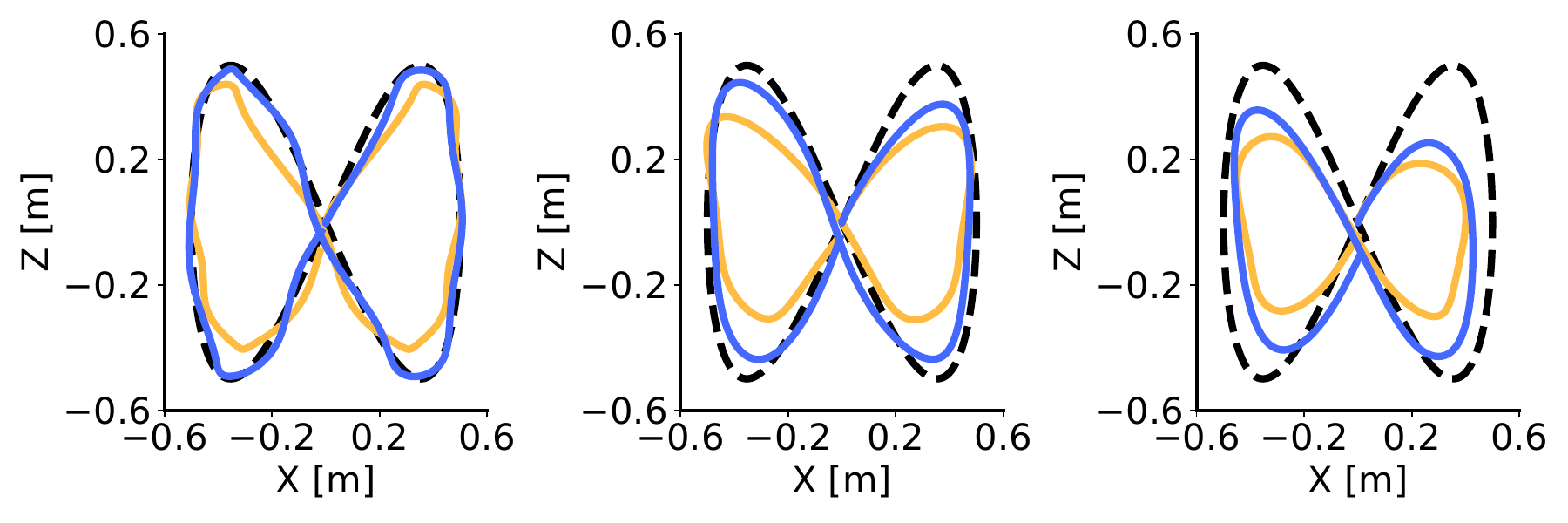}
    \vspace{-0.5cm}
    \caption{\textit{Simulation 2 —} Figure-8 trajectories tracked by Tiny FB MPC (yellow) and Tiny LB MPC (blue) at increasing angular speeds. Tiny LB MPC achieves lower RMSEs: \textbf{0.032} vs. \textbf{0.10}~m (0.5~rad/s), \textbf{0.086} vs. \textbf{0.152}~m (1~rad/s), and \textbf{0.140} vs. \textbf{0.193}~m (right).}
     \label{fig:sim}
\end{figure}
\begin{figure}[t]
     \centering
     \includegraphics[width=0.35\textwidth]{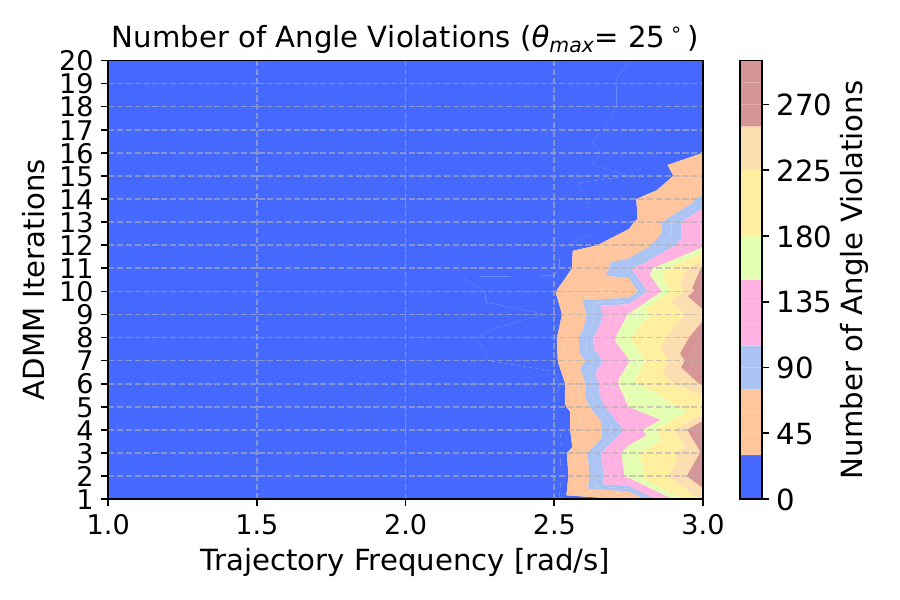}
     \vspace{-0.5cm}
     \caption{\textit{Simulation 3 —} Thrust-angle violations 
($\theta_{\max}=25^{\circ}$) across trajectory frequencies and ADMM iterations. 
Tiny~LB~MPC remains feasible with under five iterations, 
while higher frequencies reveal a trade-off between solver speed and constraint satisfaction.}
\label{fig:contour_plots}
 \end{figure}

\section{Simulations}
Our hardware implementation on the Teensy 4.0 microcontroller is limited by memory and computation. Accordingly, \textit{Simulations 1--3} evaluate two key parameters influencing safety and performance: (i) the number of GP training points, and (ii) the number of ADMM iterations (up to 5 iterations at 100~Hz on the Teensy). In \textit{Simulation~1}, we vary the number of GP data points sampled via Latin hypercubic sampling~\citep{langaker_cautious_nodate} and compare the root mean square error (RMSE) of Tiny LB MPC on a circular trajectory. As shown in Fig.~\ref{fig:rmse_boxplot}, increasing GP data points improves median RMSE and reduces variance but increases memory and inference cost, slowing the pre-computation block in Fig.~\ref{fig:system}. A trade-off of 20 data points was chosen to balance performance and rate, running pre-computation at 10~Hz. In \textit{Simulation~2}, we compare Tiny LB MPC to Tiny Flatness-Based MPC~\citep{greeff_flatness-based_2018} using the ADMM solver~\citep{nguyen_tinympc_2024} on a figure-eight trajectory under unmodelled linear drag $f_d(\mathbf{x})=\mathbf{RDR}^\top\mathbf{v}$, where $\mathbf{D}=\text{diag}[0.1,0.1,0.1]$~\citep{faessler_differential_2018}. Constraints are set to $\theta_{\text{max}}=25^\circ$ and $c_{\text{max}}=17$~m/s$^2$. Fig.~\ref{fig:sim} shows that Tiny LB MPC outperforms Tiny FB MPC by 27--68\% despite using only 20 GP points. Finally, \textit{Simulation~3} (Fig.~\ref{fig:contour_plots}) examines how the number of ADMM iterations affects thrust constraint satisfaction. For moderate trajectories, 3--5 iterations---enabling 100~Hz operation---are sufficient, whereas highly aggressive trajectories may cause constraint violations.

\section{Experiments}
\begin{figure}[t]
    \centering
    \subfigure[\textit{Experiment 1 — Circular trajectory.}]{
        \includegraphics[width=0.45\linewidth, trim={1cm 0cm 1cm 1cm}]{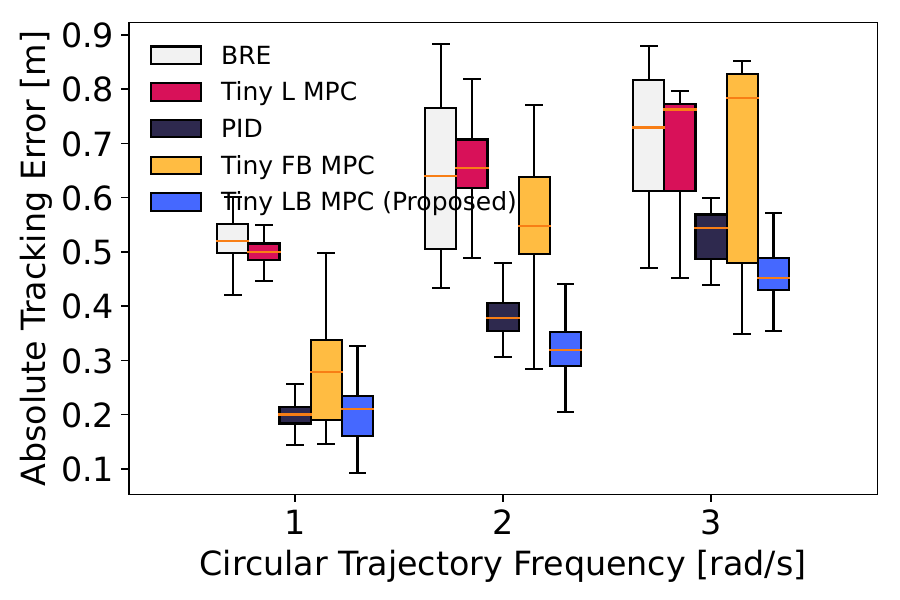}
        \label{fig:exp1}
    } 
    \hspace{0.2cm}
    \subfigure[\textit{Experiment 2 — Multiple trajectories.}]{
        \includegraphics[width=0.45\linewidth, trim={1cm 0cm 1cm 1cm}]{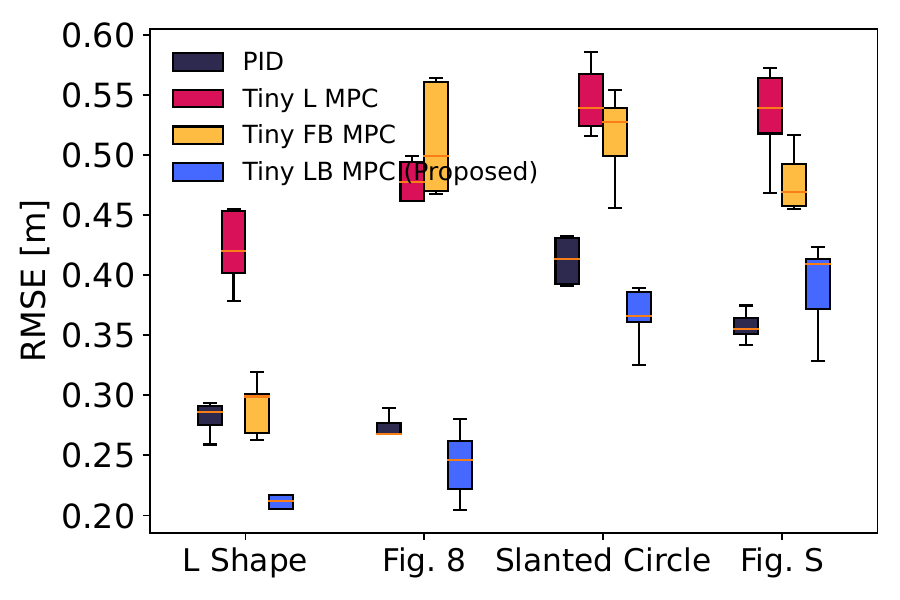}
        \label{fig:exp2}
    } 
    \caption{
    Performance comparison of embedded controllers. 
    In \textit{Experiment~1}, Tiny~LB~MPC (blue) achieves the lowest absolute tracking error on a circular trajectory, outperforming Brescianini (white), PID (black), Tiny~L~MPC (red), and Tiny~FB~MPC (yellow). 
    In \textit{Experiment~2}, RMSE over five trials for L Shape, Figure-8, Slanted Circle, and Figure S trajectories shows Tiny~LB~MPC consistently achieves the best tracking accuracy.
    }
    \label{fig:experiments}
\end{figure}

\begin{figure}[t]
    \centering
    \subfigure{
        \centering
        \includegraphics[width=\linewidth]{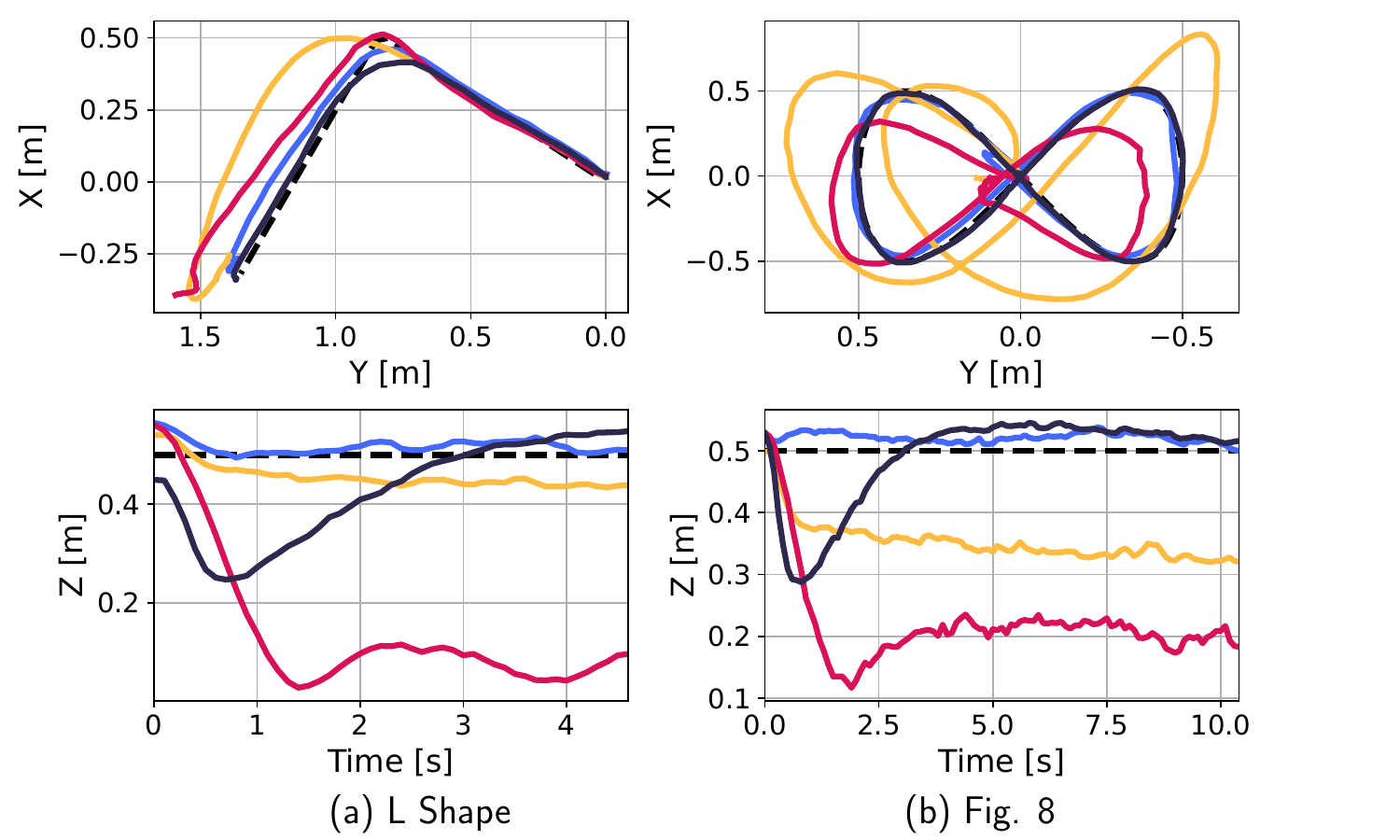}
        }
    \subfigure{
        \centering
        \includegraphics[width=\linewidth]{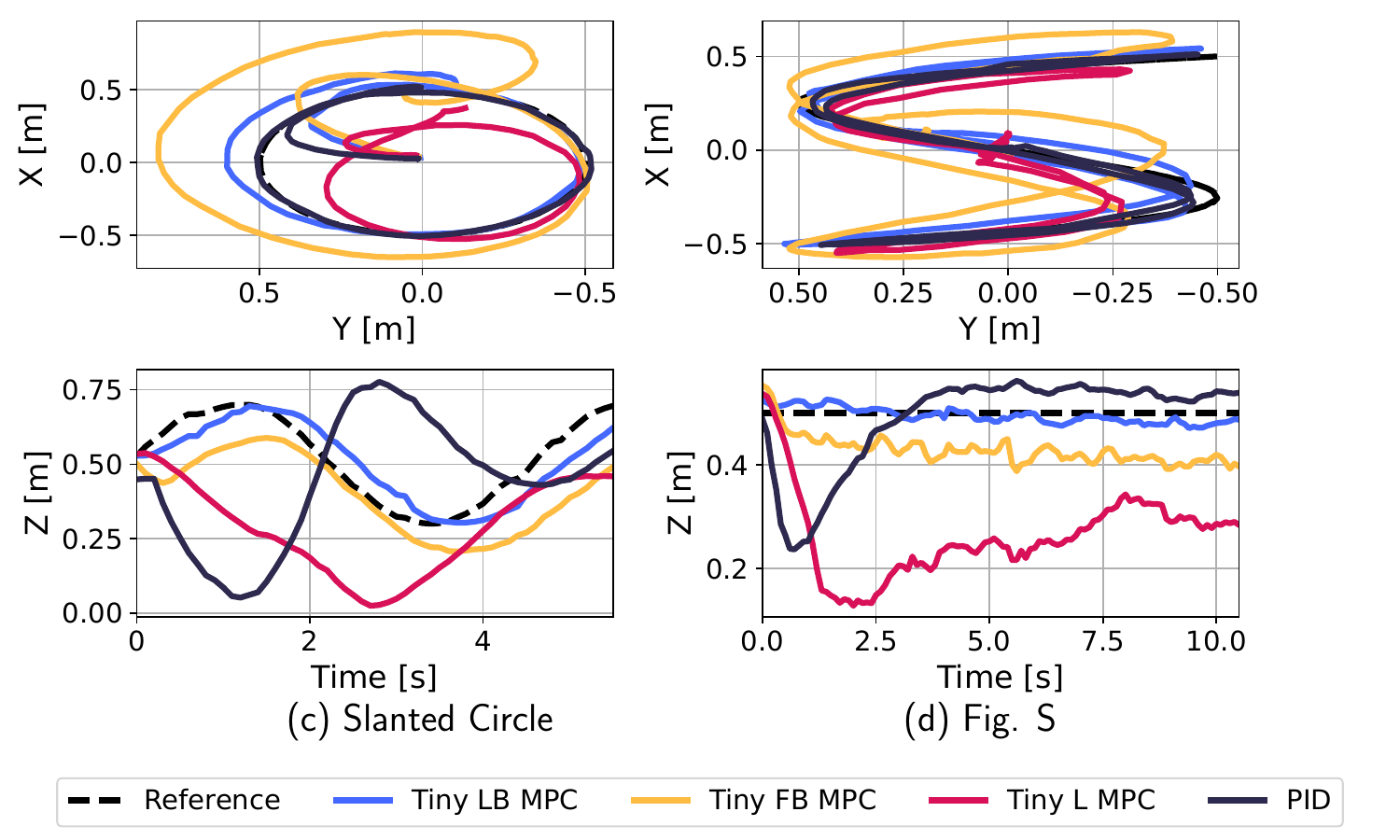}
        }
    \vspace{-0.5cm}
    \caption{Visualization of four different trajectories: (a) L Shape, (b) Figure 8, (c) Slanted Circle, and (d) Figure S, tracked by four different controllers—proposed Tiny LB MPC (blue), Tiny FB MPC (yellow), Tiny L MPC (red), and PID (black). Our method achieves the least RMSE among all the other controllers in tracking the mentioned trajectories.}
    \label{fig:exp2_2}
\end{figure}

\begin{figure}[t]
    \centering
    
    \subfigure[20 cm]{
        \includegraphics[width=0.4\columnwidth, trim={0cm 0cm 0cm 0cm}]{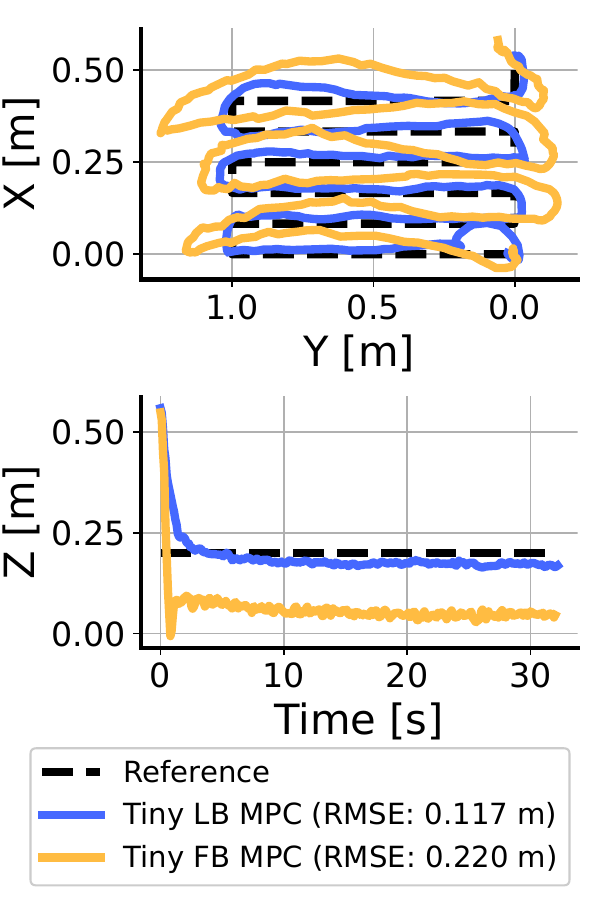}
        \label{fig:ge_1}
    } 
    \subfigure[1 m]{
        \includegraphics[width=0.4\columnwidth, trim={0cm 0cm 0cm 0cm}]{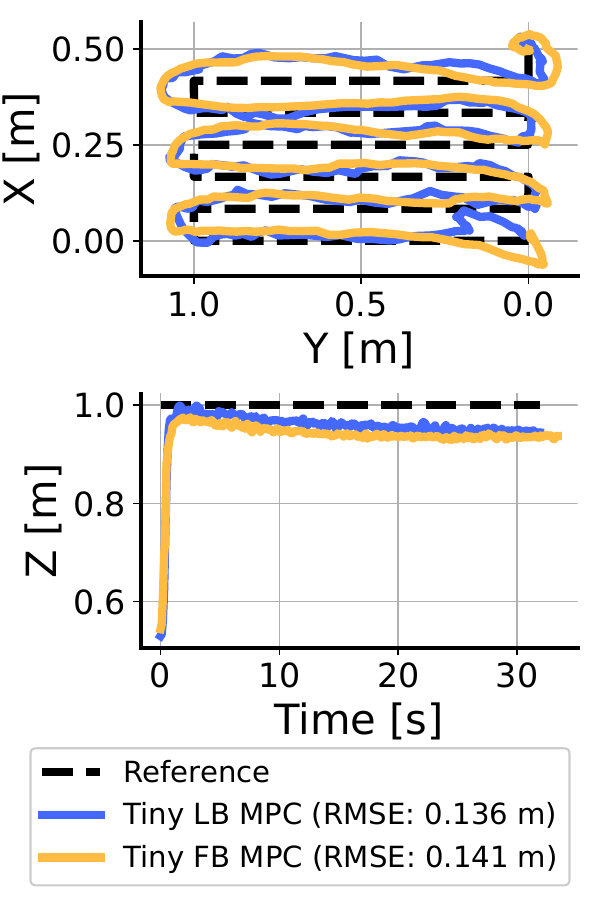}
        \label{fig:ge_2}
    }
    \vspace{-0.3cm}
    \caption{\textit{Experiment 3 —} Performance comparison of Tiny LB MPC (blue) and Tiny FB MPC (yellow) on a floor-sweeping trajectory. At 20 cm altitude (a), Tiny LB MPC compensates for ground effect, achieving an RMSE of \textbf{0.12 m} vs. \textbf{0.22 m} for Tiny FB MPC. At 1 m (b), it still performs better: \textbf{0.136 m} vs. \textbf{0.141 m}.}
    \label{fig:exp3}
\end{figure}

As shown in Fig.~\ref{fig:teensy}, we implement our controller on a modified Crazyflie~2.1 platform equipped with a custom \textit{Teensy deck}, which adds a Teensy~4.0 board~\citep{teensy40_nodate} featuring a 600~MHz Cortex-M7 MCU, 1024~KB RAM, and 2~MB flash. The deck provides an external processor for running Tiny~LB~MPC and communicates with the Crazyflie via UART at 2.25~Mbps. State estimation, including acceleration, is obtained from the Crazyflie’s onboard EKF~\citep{bitcraze_state_nodate}. We use FreeRTOS~\citep{freertos_nodate} to separate the 10~Hz pre-computation task (GP inference and matrix updates) from the 100~Hz high-rate ADMM solver, running 3--5 iterations per control cycle. The platform experiences both internal mismatch and external disturbances. Internal mismatch arises from the additional payload mass and the use of a linear thrust--PWM mapping (from the default PID controller). The GP model, trained on 20 samples from a Tiny~FB~MPC flight, shows a maximum standard deviation of 1--3~m/s\textsuperscript{2} depending on sampling. We compare our controller against Brescianini (BRE)~\citep{brescianini_nonlinear_2013}, PID, Tiny~FB~MPC, and Tiny~L~MPC~\citep{nguyen_tinympc_2024}. 
\textit{Experiment~1} evaluates performance on a circular trajectory at increasing angular frequencies. Fig.~\ref{fig:exp1} shows that Tiny~LB~MPC achieves up to 9\% improvement over PID, 36\% over Tiny~FB~MPC, 49\% over BRE, and 50\% over Tiny~L~MPC. In \textit{Experiment~2}, we compare RMSE across multiple trajectories (L~Shape, Figure~8, Slanted~Circle, and Figure~S) over five trials. As shown in Fig.~\ref{fig:exp2}, Tiny~LB~MPC consistently outperforms the other embedded MPCs that neglect model uncertainty, except for Figure~S, where its performance approaches PID. Improved performance is expected with adaptive data selection. Fig.~\ref{fig:exp2_2} illustrates sample trajectories. In \textit{Experiment~3}, we assess robustness to external disturbances such as ground effect. The Crazyflie tracks a floor-sweeping trajectory at 20~cm and 1~m altitude. As shown in Fig.~\ref{fig:ge_1}, Tiny~LB~MPC compensates for ground effect, achieving an RMSE of 0.12~m vs.\ 0.22~m for Tiny~FB~MPC. At 1~m (Fig.~\ref{fig:ge_2}), it still performs better, with RMSE 0.136~m vs.\ 0.141~m.

\section{Conclusion}

We present Tiny LB MPC deployed on a low-cost, 53~g multirotor. By combining differential flatness, a GP model, and an ADMM-based solver, it achieves 100~Hz control on the Crazyflie~2.1, outperforming prior embedded MPC methods that ignore unmodeled dynamics. Simulations and experiments show robustness to internal mismatches (mass, thrust nonlinearities) and external disturbances (drag, ground effect). Using only 20 GP points and 3--5 ADMM iterations, the controller runs efficiently with high tracking accuracy. Future work will explore data selection, variable pre-computation rate, and online adaptation to time-varying disturbances (e.g., wind).

\bibliography{ifacconf}             %

\end{document}